\documentclass[10pt,twocolumn,letterpaper]{article}

\pdfoutput=1
\usepackage{iccv}
\usepackage{times}
\usepackage{epsfig}
\usepackage{graphicx}
\usepackage{amsmath}
\usepackage{amssymb}

\usepackage{url}
\usepackage{amsfonts}
\usepackage{amsthm}

\usepackage{multirow}
\usepackage{booktabs}
\usepackage{color}
\usepackage[]{algorithm2e}

\iccvfinalcopy % *** Uncomment this line for the final submission

\def\iccvPaperID{1161} % *** Enter the ICCV Paper ID here

\begin{document}

%%%%%%%%% TITLE
\title{Interpretable Learning for Self-Driving Cars by Visualizing Causal Attention}

\author{Jinkyu Kim and John Canny\\
Computer Science Division \\University of California, Berkeley\\
Berkeley, CA 94720, USA\\
{\tt\small \{jinkyu.kim, canny\}@berkeley.edu}
}

\maketitle
%\thispagestyle{empty}

%%%%%%%%% ABSTRACT
\begin{abstract}
Deep neural perception and control networks are likely to be a key component of self-driving vehicles. These models need to be explainable - they should provide easy-to-interpret rationales for their behavior - so that passengers, insurance companies, law enforcement, developers etc., can understand what triggered a particular behavior. Here we explore the use of visual explanations. These explanations take the form of real-time highlighted regions of an image that causally influence the network's output (steering control). Our approach is two-stage. In the first stage, we use a visual attention model to train a convolution network end-to-end from images to steering angle. The attention model highlights image regions that {\em potentially influence} the network's output. Some of these are true influences, but some are spurious. We then apply a causal filtering step to determine which input regions actually influence the output. This produces more succinct visual explanations and more accurately exposes the network's behavior. We demonstrate the effectiveness of our model on three datasets totaling 16 hours of driving. We first show that training with attention does not degrade the performance of the end-to-end network. Then we show that the network causally cues on a variety of features that are used by humans while driving. 
\end{abstract}

\section{Introduction}
\begin{figure*}
    \begin{center}
        \includegraphics[width=0.95\linewidth]{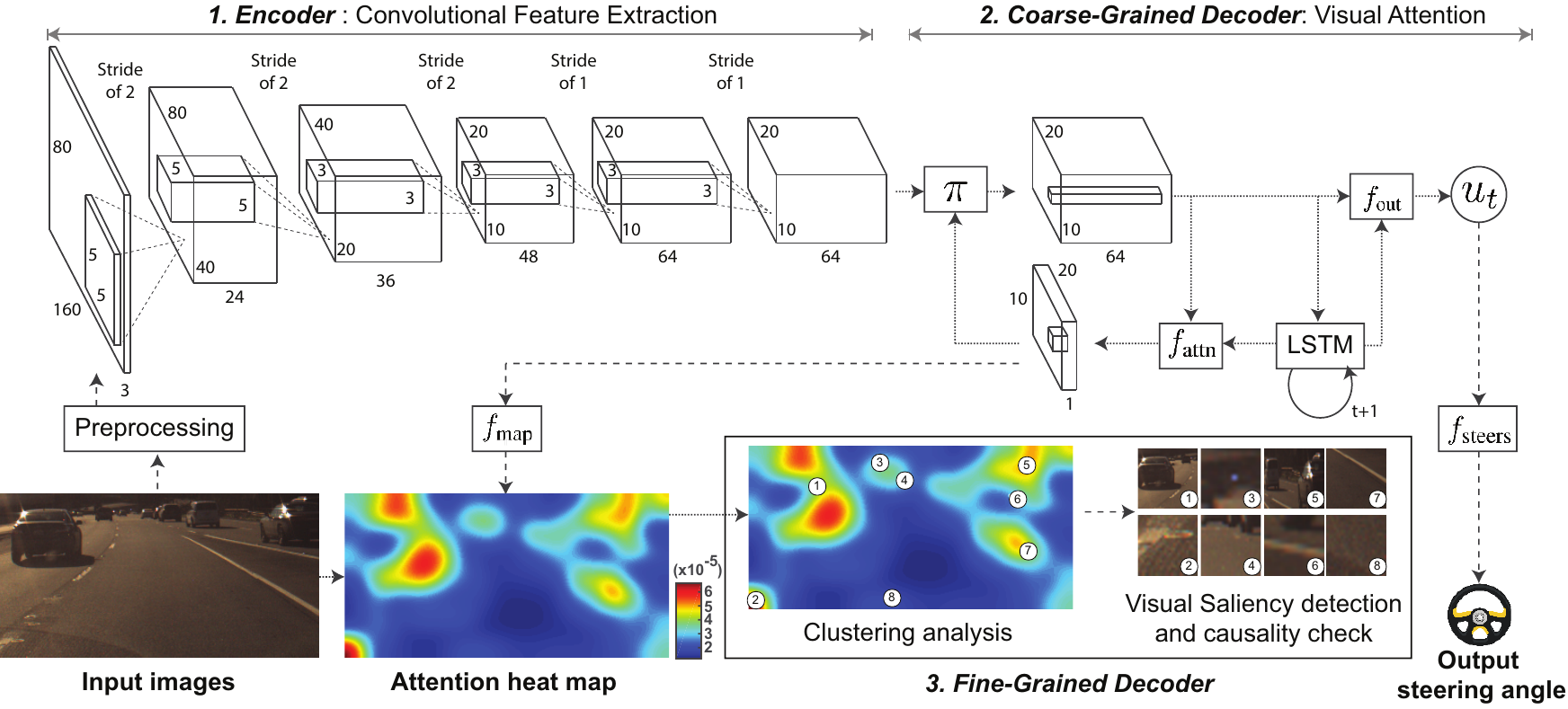}
    \end{center}
    \caption{Our model predicts steering angle commands from an input raw image stream in an end-to-end manner. In addition, our model generates a heat map of attention, which can visualize where and what the model sees. To this end, we first encode images with a CNN and decode this feature into a heat map of attention, which is also used to control a vehicle. We test its causality by scrutinizing each cluster of attention blobs and produce a refined attention heat map of causal visual saliency.}
    \label{fig:overview}
\end{figure*}

Self-driving vehicle control has made dramatic progress in the last several years, and many auto vendors have pledged large-scale commercialization in a 2-3 year time frame. These controllers use a variety of approaches but recent successes \cite{bojarski2016end} suggests that neural networks will be widely used in self-driving vehicles. But neural networks are notoriously cryptic - both network architecture and hidden layer activations may have no obvious relation to the function being estimated by the network. An exception to the rule is visual attention networks \cite{xu2015show,sharma2015action,chen2017brain}. These networks provide spatial attention maps - areas of the image that the network attends to - that can be displayed in a way that is easy for users to interpret. They provide their attention maps instantly on images that are input to the network, and in this case on the stream of images from automobile video. As we show from our examples later, visual attention maps lie over image areas that have intuitive influence on the vehicle's control signal.

But attention maps are only part of the story. Attention is a mechanism for filtering out non-salient image content. But attention networks need to find all {\em potentially} salient image areas and pass them to the main recognition network (a CNN here) for a final verdict. For instance, the attention network will attend to trees and bushes in areas of an image where road signs commonly occur. Just as a human will use peripheral vision to determine that "there is something there", and then visually fixate on the item to determine what it actually is. We therefore post-process the attention network's output, clustering it into attention "blobs" and then mask (set the attention weights to zero) each blob to determine the effect on the end-to-end network output. Blobs that have an causal effect on network output are retained while those that do not are removed from the visual map presented to the user. 

Figure~\ref{fig:overview} shows an overview of our model. Our approach can be divided into three steps: (1) Encoder: convolutional  feature extraction, (2) Coarse-grained decoder by visual attention mechanism, and (3) Fine-grained decoder: causal visual saliency detection and refinement of attention map. Our contributions are as follows: 

\vspace{-1mm}
\begin{itemize}
\item We show that visual attention heat maps are suitable "explanations" for the behavior of a deep neural vehicle controller, and do not degrade control accuracy.\vspace{-3mm}
\item We show that attention maps comprise "blobs" that can be segmented and filtered to produce simpler and more accurate maps of visual saliency.\vspace{-3mm}
\item We demonstrate the effectiveness of using our model with three large real-world driving datasets that contain over 1,200,000 video frames (\textit{approx.} 16 hours).\vspace{-3mm}
\item We illustrate typical spurious attention sources in driving video and quantify the reduction in explanation complexity from causal filtering.\vspace{-1mm}
\end{itemize}

\section{Related Works}
\subsection{End-to-End Learning for Self-driving Cars}
Self-driving vehicle controllers can be classified as: mediated perception approaches and end-to-end learning approaches. The mediated perception approach depends on recognizing human-designated features (\ie, lane markings and cars) in a controller  with if-then-else rules. Some examples include Urmson~\etal~\cite{urmson2008autonomous}, Buehler~\etal~\cite{buehler2009darpa}, and Levinson~\etal~\cite{levinson2011towards}. 

Recently there is growing interest in end-to-end learning vehicle control. Most of these approaches learn a controller by supervised regression to recordings from human drivers. The training data comprise video from one or more vehicle cameras, and the
control outputs (steeting and possible acceleration and braking) from the driver. 
ALVINN (Autonomous Land Vehicle In a Neural Network)~\cite{pomerleau1989alvinn} was the first attempt to use neural network for directly mapping images to navigate the direction of the vehicle. More recently Bojarski \etal~\cite{bojarski2016end} demonstrated good performance with convolutional neural networks (CNNs) to directly map images from a front-view camera to steering controls. Xu~\etal~\cite{xu2016end} proposed an end-to-end egomotion prediction approach that takes raw pixels and prior vehicle state signals as inputs and predicts several a sequence of discretized actions (\ie, straight, stop, left-turn, and right-turn). These models show good performance but their behavior is opaque and uninterpretable.

An intermediate approach was explored in Chen~\etal~\cite{chen2015deepdriving} who defined human-interpretable intermediate features such as the curvature of lane, distances to neighboring lanes, and distances from the front-located vehicles. A CNN is trained to produce these features, and a simple controller maps them to steering angle. They also generated deconvolution maps to show image areas that affected network output. However, there were several difficulties with that work: (i) use of the intermediate layer caused significant degradation (40\% or more) of control accuracy (ii) the intermediate feature descriptors provide a limited and ad-hoc vocabulary for explanations and (iii) the authors noted the presence of spurious input features but there was no attempt to remove them. By contrast, our work shows that state-of-the-art driving models can be made interpretable without sacrificing accuracy, that attention models provide more robust image annotation, and causal analysis further improves explanation saliency. 

\subsection{Visual Explanation}
In a landmark work, Zeiler and Fergus~\cite{zeiler2014visualizing} used "deconvolution" to visualize layer activations of convolutional networks. LeCun~\etal~\cite{lecun2015deep} provides textual explanations of images as automatically-generated captions. Building on this work, Bojarski~\etal~\cite{bojarski2016visualbackprop} developed a richer notion of "contribution" of a pixel to the output. However a difficulty with deconvolution-style approaches is the lack of formal measures of how the network output is affected by spatially-extended features (rather than pixels). Attention-based approaches like ours directly extract areas of the image that {\em did not affect} network output (because they were masked out by the attention model), and causal filtering further removes spurious image areas. 
Hendricks~\etal~\cite{hendricks2016generating} trains a deep network to generate species specific explanation without explicitly identifying semantic features. Also, Justin Johnson~\etal~\cite{johnson2016densecap} proposes DenseCap which uses fully convolutional localization networks for dense captioning, their paper achieves both localizing objects and describing salient regions in images using natural langauge. In reinforcement learning, Zrihem~\etal~\cite{zahavy2016graying} proposes a visualization method to interpret the agent’s action by describing Markov Decision Process model as a directed graph on a t-SNE map. 

\section{Method}
\subsection{Preprocessing}~\label{ss:preprocessing}
Our model predicts continuous steering angle commands from input raw pixels in an end-to-end manner. As discussed by Bojarski~\etal~\cite{bojarski2016end}, our model predicts the inverse turning radius $\hat{u}_t$ (= $r_t^{-1}$, where $r_t$ is the turning radius) at every timestep $t$ instead of steering angle commands, which depends on the vehicle's steering geometry and also result in numerical instability when predicting near zero steering angle commands. The relationship between the inverse turning radius $u_t$ and the steering angle command $\theta_t$ can be approximated by Ackermann steering geometry~\cite{rajamani2011vehicle} as follows: 
\begin{equation}
\theta_{t} = f_{\textnormal{steers}}(u_t) = u_t d_\textnormal{w} K_\textnormal{s} (1+K_\textnormal{slip}{v_t}^2)
\end{equation}
where $\theta_{t}$ in degrees and $v_t$ (m/s) is a steering angle and a velocity at time $t$, respectively. $K_\textnormal{s}$, $K_\textnormal{slip}$, and $d_\textnormal{w}$ are vehicle-specific parameters. $K_\textnormal{s}$ is a steering ratio between the turn of the steering and the turn of the wheels. $K_\textnormal{slip}$ represents the relative motion between a wheel and the surface of road. $d_\textnormal{w}$ is the length between the front and rear wheels. Our model therefore needs two measurements for training: timestamped vehicle's speed and steering angle commands.

To reduce computational cost, each raw input image is down-sampled and resized to 80$\times$160$\times$3 with nearest-neighbor scaling algorithm. For images with different raw aspect ratios, we cropped the height to match the ratio before down-sampling. We also normalized pixel values to $[0,1]$ in HSV colorspace.

We utilize a single exponential smoothing method~\cite{hyndman2008forecasting} to reduce the effect of human factors-related performance variation and the effect of measurement noise. Formally, given a smoothing factor $0\leq\alpha_{\textnormal{s}}\leq1$, the simple exponential smoothing method is defined as follows:
\begin{equation}
\begin{pmatrix} \hat{\theta}_t  \\ \hat{v}_t \end{pmatrix} = \alpha_{s} \begin{pmatrix} \theta_t  \\ v_t \end{pmatrix} + (1-\alpha_{s})\begin{pmatrix} \hat{\theta}_{t-1}  \\ \hat{v}_{t-1} \end{pmatrix}
\end{equation}
where $\hat{\theta}_t$ and $\hat{v}_t$ are the smoothed time-series of $\theta_t$ and $v_t$, respectively. Note that they are same as the original time-series when $\alpha_{\textnormal{s}}=1$, while values of $\alpha_{\textnormal{s}}$ closer to zero have a greater smoothing effect and are less responsive to recent changes. The effect of applying smoothing methods is summarized in Section~\ref{ss:smoothing}.

\subsection{Encoder: Convolutional Feature Extraction}\label{ss:feature}
We use a convolutional neural network to extract a set of encoded visual feature vector, which we refer to as a convolutional feature cube ${\bf{x}_t}$. Each feature vectors may contain high-level object descriptions that allow the attention model to selectively pay attention to certain parts of an input image by choosing a subset of feature vectors.

As depicted in Figure~\ref{fig:overview}, we use a 5-layered convolution network that is utilized by Bojarski~\etal~\cite{bojarski2016end} to learn a model for self-driving cars. As discussed by Lee~\etal~\cite{lee2009convolutional}, we omit max-pooling layers to prevent spatial locational information loss as the strongest activation propagates through the model. We collect a three-dimensional convolutional feature cube ${\bf{x}_t}$ from the last layer by pushing the preprocessed image through the model, and the output feature cube will be used as an input of the LSTM layers, which we will explain in Section~\ref{ss:LSTM}. Using this convolutional feature cube from the last layer has advantages in generating high-level object descriptions, thus increasing interpretability and reducing computational burdens for a real-time system.

Formally, a convolutional feature cube of size $W$$\times$$H$$\times$$D$ is created at each timestep $t$ from the last convolutional layer. We then collect ${\bf{x}}_{t}$, a set of $L=W\times H$ vectors, each of which is a $D$-dimensional feature slice for different spatial parts of the given input.
\begin{equation}
{\bf{x}}_{t} = \{ x_{t,1}, x_{t,2},\ldots, x_{t,L} \}
\end{equation}
where $x_{t,i}\in{\cal{R}}^{D}$ for $i\in\{1,2,\ldots,L\}$. This allows us to focus selectively on different spatial parts of the given image by choosing a subset of these $L$ feature vectors.

\subsection{Coarse-Grained Decoder: Visual Attention}\label{ss:LSTM}
The goal of soft deterministic attention mechanism $\pi(\{x_{t,1}, x_{t,2}, \dots, x_{t,L}\})$ is to search for a good context vector $y_t$, which is defined as a combination of convolutional feature vectors $x_{t,i}$, while producing better prediction accuracy. We utilize a deterministic soft attention mechanism that is trainable by standard back-propagation methods, which thus has advantages over a hard stochastic attention mechanism that requires reinforcement learning. Our model feeds $\alpha$ weighted context $y_t$ to the system as discuss by several works~\cite{sharma2015action, xu2015show}:
\begin{equation}
\begin{split}
y_t & =  f_\textnormal{flatten}(\pi(\{\alpha_{t,i}\}, \{x_{t,i}\})) \\
 & = f_\textnormal{flatten}(\{\alpha_{t,i}x_{t,i}\})
\end{split}
\end{equation}
where $i=\{1,2,\dots,L\}$. $\alpha_{t,i}$ is a scalar attention weight value associated with a certain grid of input image in such that $\sum_{i}{\alpha_{t,i}}=1$. These attention weights can be interpreted as the probability over $L$ convolutional feature vectors that the location $i$ is the important part to produce better estimation accuracy.  $f_\textnormal{flatten}$ is a flattening function. $y_t$ is thus $D$$\times$$L$-dimensional vector that contains convolutional feature vectors weighted by attention weights. Note that, our attention mechanism $\pi(\{\alpha_{t,i}\}, \{x_{t,i}\})$ is different from the previous works~\cite{sharma2015action, xu2015show}, which use the $\alpha$ weighted average context $y_t=\sum_{i=1}^{L} \alpha_{t,i}x_{t,i}$. We observed that this change significantly improves overall prediction accuracy. The performance comparison is explained in Section~\ref{ss:quant}.

As we summarize in Figure~\ref{fig:overview}, we use a long short-term memory (LSTM) network~\cite{hochreiter1997long} that predicts the inverse turning radius $\hat{u}_t$ and generates attention weights $\{\alpha_{t,i}\}$ at each timestep $t$ conditioned on the previous hidden state $h_t$ and a current convolutional feature cube ${\bf{x_t}}$. More formally, let us assume a hidden layer $f_{\textnormal{attn}}({x_{t,i}},h_{t-1})$ conditioned on the previous hidden state $h_{t-1}$ and the current feature vectors $\{x_{t,i}\}$. The attention weight $\{\alpha_{t,i}\}$ for each spatial location $i$ is then computed by multinomial logistic regression (\ie, softmax regression) function as follows:
\begin{equation}
\alpha_{t,i} = \frac{\exp(f_{\textnormal{attn}}(x_{t,i},h_{t-1}))}{\sum_{j=1}^{L}\exp(f_{\textnormal{attn}}(x_{t,j},h_{t-1}))}
\end{equation}

Our network also predicts inverse turning radius $\hat{u}_t$ as an output with additional hidden layer $f_{\textnormal{out}}(y_t, h_t)$ conditioned on the current hidden state $h_t$ and $\alpha$ weighted context $y_t$.

To initialize memory state $c_t$ and hidden state $h_t$ of LSTM network, we follow Xu~\etal~\cite{xu2015show} by averaging of the feature slices $x_{0,i}$ at initial time fed through two additional hidden layers: $f_{\textnormal{init},c}$ and $f_{\textnormal{init},h}$.
\begin{equation}
c_{0} = f_{\textnormal{init},c}\left(\frac{1}{L}\sum^{L}_{i=1}x_{0,i}\right),~~
h_{0} = f_{\textnormal{init},h}\left(\frac{1}{L}\sum^{L}_{i=1}x_{0,i}\right)
\end{equation}

As discussed by Xu~\etal~\cite{xu2015show}, doubly stochastic regularization can encourage the attention model to at different parts of the image. At time $t$, our attention model predicts a scalar $\beta$=$\textit{sigm}(f_{\beta}(h_{t-1}))$ with an additional hidden layer $f_{\beta}$ conditioned on the previous hidden state $h_{t-1}$ such that 
\begin{equation}
y_t = \textit{sigm}(f_{\beta}(h_{t-1})) f_\textnormal{flatten}(\{\alpha_{t,i} x_{t,i} \})
\end{equation}

We use the following penalized loss function $\mathcal{L}_1$: 
\begin{equation}
\mathcal{L}_1(u_t, \hat{u}_t) = \sum_{t=1}^{T}|u_{t} - \hat{u}_{t}|+\lambda\sum_{i=1}^{L}\left(1-\sum_{t=1}^{T}\alpha_{t,i}\right)
\end{equation}
where $T$ is the length of time steps, and $\lambda$ is a penalty coefficient that encourages the attention model to see different parts of the image at each time frame. Section~\ref{ss:attention} describes the effect of using regularization.

\begin{figure*}[!t]
    \begin{center}
        \includegraphics[width=0.95\linewidth]{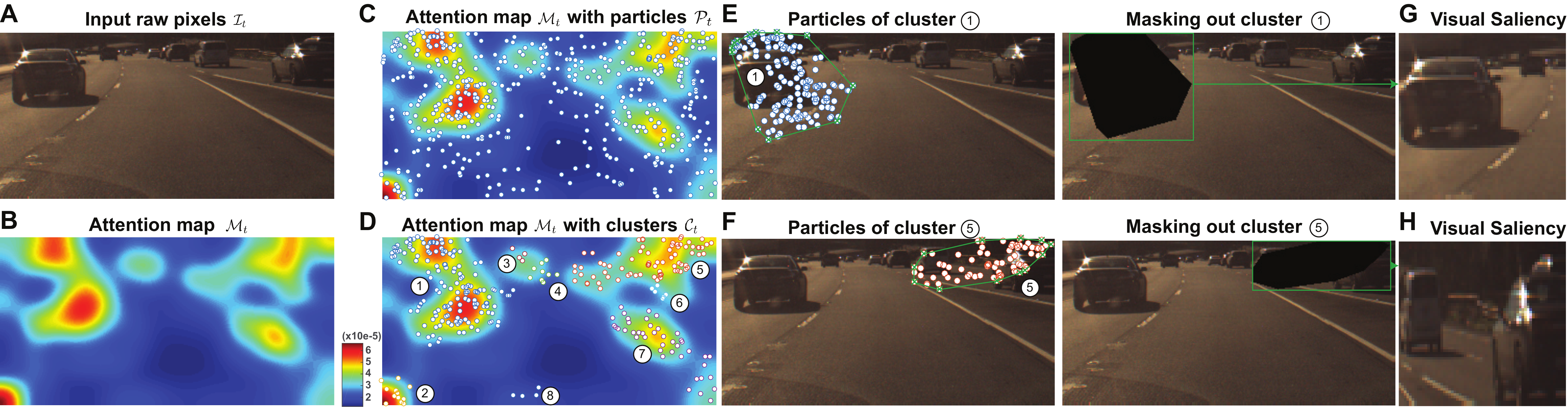}
    \end{center}
    \caption{Overview of our fine-grained decoder. Given an input raw pixels $\mathcal{I}_t$ (A), we compute an attention map $\mathcal{M}_t$ with a function $ f_{\textnormal{map}}$ (B). (C) We randomly sample 3D $N=500$ particles over the attention map, and (D) we apply a density-based clustering algorithm (DBSCAN~\cite{ester1996density}) to find a local visual saliency by grouping particles into clusters. (E, F) For each cluster $c\in\mathcal{C}$, we compute a convex hull $\mathcal{H}(c)$ to define its region, and mask out the visual saliency to see causal effects on prediction accuracy  (see E, F for clusters 1 and 5, respectively). (G, H) Warped visual saliencies for clusters 1 and 5, respectively.}
    \label{fig:finegrained}
\end{figure*}

\subsection{Fine-Grained Decoder: Causality Test}\label{ss:detection}
The last step of our pipeline is a fine-grained decoder, in which we refine a map of attention and detect local visual saliencies. Though an attention map from our coarse-grained decoder provides probability of importance over a 2D image space, our model needs to determine specific regions that cause a causal effect on prediction performance. To this end, we assess a decrease in performance when a local visual saliency on an input raw image is masked out. 

We first collect a consecutive set of attention weights  \{$\alpha_{t, i}$\} and input raw images \{$\mathcal{I}_t$\} for a user-specified $T$ timesteps. We then create a map of attention, which we refer $\mathcal{M}_t$ as defined: $\mathcal{M}_{t}=f_{\textnormal{map}}(\{\alpha_{{t},i}\})$. Our 5-layer convolutional neural network uses a stack of $5\times 5$ and $3\times 3$ filters without any pooling layer, and therefore the input image of size $80\times160$ is processed to produce the output feature cube of size $10\times20\times64$, while preserving its aspect ratio. Thus, we use $f_{\textnormal{map}}(\{\alpha_{{t},i}\})$ as up-sampling function by the factor of eight followed by Gaussian filtering~\cite{burt1983laplacian} as discussed in \cite{xu2015show} (see Figure~\ref{fig:finegrained} (A,B)). 

To extract a local visual saliency, we first randomly sample 2D $N$ particles with replacement over an input raw image conditioned on the attention map $\mathcal{M}_{t}$.  Note that, we also use time-axis as the third dimension to consider temporal features of visual saliencies. We thus store spatio-temporal 3D particles $\mathcal{P} \gets \mathcal{P} \cup \{\mathcal{P}_t, t\}$ (see Figure~\ref{fig:finegrained} (C)). 

We then apply a clustering algorithm to find a local visual saliency by grouping 3D particles $\mathcal{P}$ into clusters $\{\mathcal{C}\}$ (see Figure~\ref{fig:finegrained} (D)). In our experiment, we use DBSCAN~\cite{ester1996density}, a density-based clustering algorithm that has advantages to deal with a noisy dataset  because they group particles together that are closely packed, while marking particles as outliers that lie alone in low-density regions. For points of each cluster $c$ and each time frame $t$, we compute a convex hull $\mathcal{H}(c)$ to find a local region of each visual saliency detected (see Figure~\ref{fig:finegrained} (E, F)). 

For points of each cluster $c$ and each time frame $t$, we iteratively measure a decrease of prediction performance with an input image which we mask out a local visual saliency. We compute a convex hull $\mathcal{H}(c)$ to find a local, and mask out each visual saliency by assigning zero values for all pixels lying inside each convex hull. Each causal visual saliency is generated by warping into a fixed spatial resolution (=64$\times$64) as shown in Figure~\ref{fig:finegrained} (G, H).

\section{Result and Discussion}
\subsection{Datasets}
As explained in Table~\ref{Table:dataset}, we obtain two large-scale datasets that contain over 1,200,000 frames ($\approx$16 hours) collected from Comma.ai~\cite{commaai}, Udacity~\cite{udacity}, and Hyundai Center of Excellence in Integrated Vehicle Safety Systems and Control (HCE) under a research contract. These three datasets acquired contain video clips captured by a single front-view camera mounted behind the windshield of the vehicle. Alongside the video data, a set of time-stamped sensor measurement is contained, such as vehicle's velocity, acceleration, steering angle, GPS location and gyroscope angles. Thus, these datasets are ideal for self-driving studies. Note that, for sensor logs unsynced with the time-stamps of video data, we use the estimates of the interpolated measurements. Videos are mostly captured during highway driving in clear weather on daytime, and there included driving on other road types, such as residential roads (with and without lane markings), and contains the whole driver's activities such as staying in a lane and switching lanes. Note also that, we exclude frames when the vehicle stops which happens when $\hat{v}_t<$1 m/s.

\begin{figure*}[t]
    \begin{center}
        \includegraphics[width=0.95\linewidth]{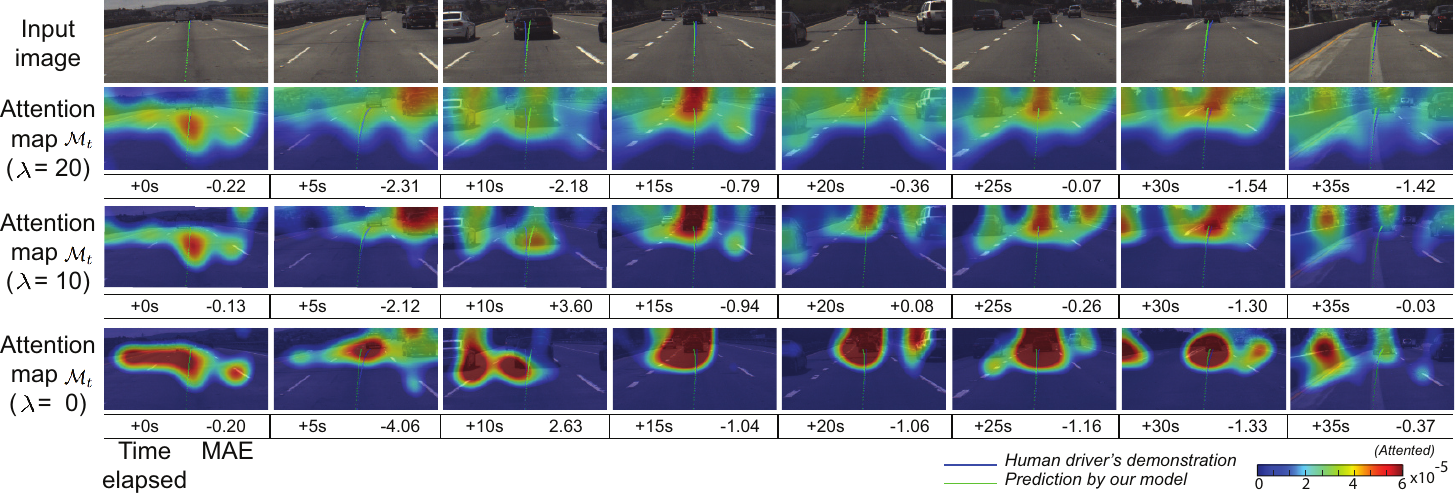}
    \end{center}
    \caption{Attention maps over time. Unseen consecutive input image frames are sampled at every 5 seconds (see from left to right). (Top) Input raw images with human driver’s demonstrated curvature of path (blue line) and predicted curvature of path (green line). (From the bottom) We illustrate attention maps with three different regularization penalty coefficients $\lambda\in\{0, 10, 20\}$.  Each attention map is overlaid by an input raw image and color-coded. Red parts indicate where the model pays attention. \textit{Data}: Comma.ai~\cite{commaai}}
    \label{fig:attention}
\end{figure*}

\subsection{Training and Evaluation Details}
To obtain a convolutional feature cube ${\bf{x}}_t$, we train the 5-layer CNNs explained in Section~\ref{ss:feature} by using additional 5-layer fully connected layers (\ie, \# hidden variables: 1164, 100, 50, and 10, respectively), of which output predicts the measured inverse turning radius $u_{t}$. Incidentally, instead of using addition fully-connected layers, we could also obtain a convolutional feature cube ${\bf{x}}_t$ by training from scratch with the whole network. In our experiment, we obtain the 10$\times$20$\times$64-dimensional convolutional feature cube, which is then flattened to 200$\times$64 and is fed through the coarse-grained decoder. Other recent types of more recent expressive networks may give a performance boost over our CNN configuration. However, exploration of other convolutional architectures would be out of our scope.

We experiment with various numbers of LSTM layers (1 to 5) of the soft deterministic visual attention model but did not observe any significant improvements in model performance. Unless otherwise stated, we use a single LSTM layer in this experiment. For training, we use Adam optimization algorithm~\cite{kingma2014adam} and also use dropout~\cite{srivastava2014dropout} of 0.5 at hidden state connections and Xavier initialization~\cite{glorot2010understanding}. We randomly sample a mini-batch of size 128, each of batch contains a set Consecutive frames of length $T=20$. Our model took less than 24 hours to train on a single NVIDIA Titan X Pascal GPU. Our implementation is based on Tensorflow~\cite{abadi2015tensorflow} and code will be publicly available upon publication.

\begin{table}[!tpb]
    \begin{center}
		\begin{tabular}{@{}llll@{}} \toprule
    		& \multicolumn{3}{c}{Dataset} \\
        	\cmidrule{2-4}
        	& Comma.ai~\cite{commaai} & HCE & Udacity~\cite{udacity} \\ \midrule
        	\# frame 	& 522,434& 80,180 & 650,690 \\
            FPS 		& 20Hz & 20Hz/30Hz & 20Hz\\
            Hours 		& $\approx$ 7 hrs & $\approx$ 1 hr & $\approx$ 8 hrs\\
        	Condition 	& Highway/Urban & Highway & Urban \\
            Location 	& CA, USA    & CA, USA & CA, USA  \\
            Lighting    & Day/Night  & Day & Day \\
        	\bottomrule
		\end{tabular}
        \caption{Dataset details. Over 16 hours ($>$1,200,000 video frames) of driving dataset that contains a front-view video frames and corresponding time-stamped measurements of vehicle dynamics. The data is collected from two public data sources, Comma.ai~\cite{commaai} and Udacity~\cite{udacity}, and Hyundai Center of Excellence in Vehicle Dynamic Systems and Control (HCE).}
        \label{Table:dataset}
    \end{center}
\end{table}

Two datasets (Comma.ai~\cite{commaai} and HCE) we used were available with images captured by a single front-view camera. This makes it hard to use the data augmentation technique proposed by Bojarski~\etal~\cite{bojarski2016end}, which generated images with artificial shifts and rotations by using two additional off-center images (left-view and right-view) captured by the same vehicle. Data augmentation may give a performance boost, but we report performance without data augmentation.

\subsection{Effect of Choosing Penalty Coefficient $\lambda $}\label{ss:attention}
Our model provides a better way to understand the rationale of the model’s decision by visualizing where and what the model sees to control a vehicle. Figure~\ref{fig:attention} shows a consecutive input raw images (with sampling period of 5 seconds) and their corresponding attention maps (\ie, $\mathcal{M}_{t}= f_{\textnormal{map}}(\{\alpha_{{t},i}\})$).  We also experiment with three different penalty coefficients $\lambda\in\{0,10,20\}$, where the model is encouraged to pay attention to wider parts of the image (see differences between the bottom 3 rows in Figure~\ref{fig:attention} ) as we have larger $\lambda$. For better visualization, an attention map is overlaid by an input raw image and color-coded; for example, red parts represent where the model pays attention. For quantitative analysis, prediction performance in terms of mean absolute error  (MAE) is explained on the bottom of each figure. We observe that our model is indeed able to pay attention on road elements, such as lane markings, guardrails, and vehicles ahead, which is essential for human to drive.

\begin{figure}[!t]
    \begin{center}
        \includegraphics[width=0.7\linewidth]{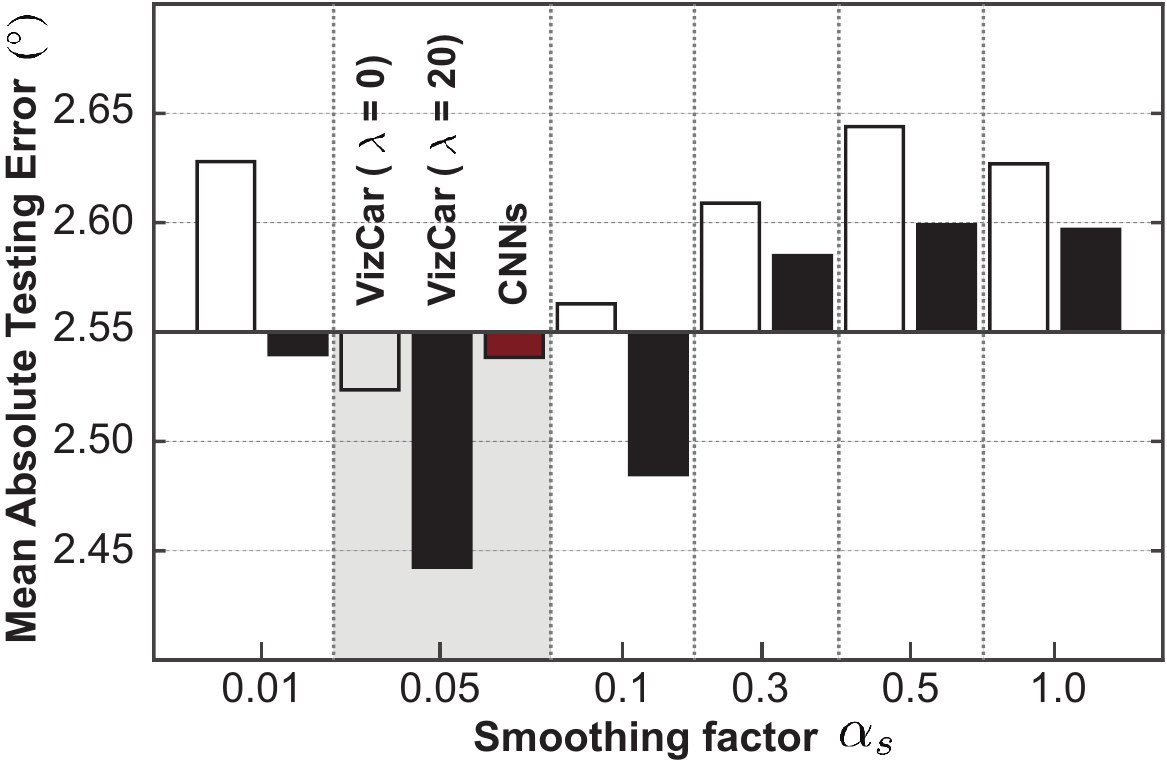}
    \end{center}
    \caption{Effect of applying a single exponential smoothing method over various smoothing factors from 0.1 to 1.0. We use two different penalty coefficients $\lambda\in\{0,20\}$. With setting $\alpha_{s}=0.05$, our model performs the best. \textit{Data}: Comma.ai~\cite{commaai}}
    \label{fig:smoothing}
\end{figure}

\subsection{Effect of Varying Smoothing Factors}~\label{ss:smoothing}
Recall from Section~\ref{ss:preprocessing} that the single exponential smoothing method~\cite{hyndman2008forecasting} is used to reduce the effect of human factors variation and the effect of measurement noise for two sensor inputs: steering angle and velocity. A robust model for autonomous vehicles would yield consistent performance, even when some measurements are noisy. Figure~\ref{fig:smoothing} shows the prediction performance in terms of mean absolute error (MAE) on a comma.ai testing data set. Various smoothing factors $\alpha_{s}\in\{0.01, 0.05, 0.1, 0.3, 0.5, 1.0\}$ are used to assess the effect of using smoothing methods. With setting $\alpha_{s}$=0.05, our model for the task of steering estimation performs the best. Unless otherwise stated, we will use $\alpha_{s}$ as 0.05.

\begin{figure*}[!t]
    \begin{center}
        \includegraphics[width=0.85\linewidth]{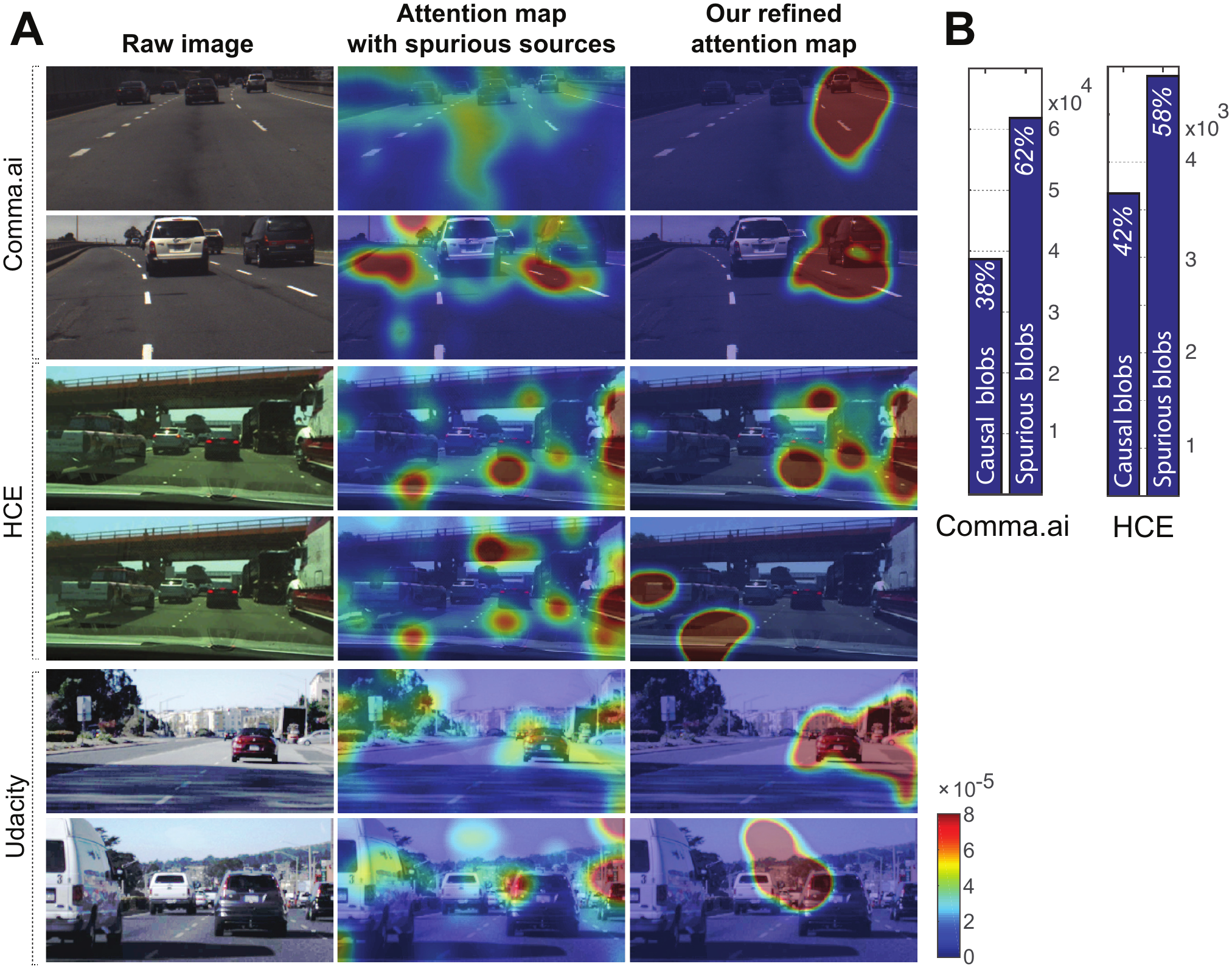}
    \end{center}
    \caption{(A) We illustrate examples of (left) raw input images, their (middle) visual attention heat maps with spurious attention sources, and (right) our attention heat maps by filtering out spurious blobs to produce simpler and more accurate attention maps. (B) To measure how much the causal filtering is simplifying attention clusters, we quantify the number of attention blobs before and after causal filtering.}
    \label{fig:causality}
\end{figure*}

\subsection{Quantitative Analysis}~\label{ss:quant}
In Table~\ref{Table:performance}, we compare the prediction performance with alternatives in terms of MAE.  We implement alternatives that include the work by Bojarski~\etal~\cite{bojarski2016end}, which used an identical base CNN and a fully-connected network (FCN) without attention. To see the contribution of LSTMs, we also test a CNN and LSTM, which is identical to ours but does not use the attention mechanism. For our model, we test with three different values of penalty coefficients $\lambda\in\{0,10,20\}$. 

Our model shows competitive prediction performance than alternatives. Our model shows 1.18--4.15 in terms of MAE on testing dataset. This confirms that incorporation of attention does not degrade control accuracy. The average run-time for our model and alternatives took less than a day to train each dataset. 

\subsection{Effect of Causal Visual Saliencies}
Recall from Section~\ref{ss:detection}, we post-process the attention network’s output by clustering it into attention blobs and filtering if they have an causal effect on network output. Figure~\ref{fig:causality} (A) shows typical examples of an input raw image, an attention networks’s output with spurious attention sources, and our refined attention heat map. We observe our model can produce a simpler and more accurate map of visual saliency by filtering out spurious attention blobs. In our experiment, 62\% and 58\% out of all attention blobs are indeed spurious attention sources on Comma.ai~\cite{commaai} and HCE datasets (see Figure~\ref{fig:causality} (B)). 

\section{Conclusion}
We described an interpretable visualization for deep self-driving vehicle controllers. It uses a visual attention model augmented with an additional layer of causal filtering. We tested with three large-scale real driving datasets that contain over 16 hours of video frames. We showed that (i) incorporation  of attention does not degrade control accuracy compared to an identical base CNN without attention (ii) raw attention highlights interpretable features in the image and (iii) causal filtering achieves a useful reduction in explanation complexity by removing features which do not significantly affect the output. 

\begin{table}[h]
	\begin{center}
    	\begin{tabular}{@{}llcc@{}} \toprule
        	\multirow{2}{*}{Dataset} & \multirow{2}{*}{Model} & \multicolumn{2}{c}{MAE in degree [SD]} \\
            & & Training & Testing\\ \midrule
        	\multirow{6}{*}{Comma.ai~\cite{commaai}} & CNN+FCN~\cite{bojarski2016end} & .421 [0.82] & 2.54 [3.19] \\ \cmidrule{2-4}
            & CNN+LSTM & .488 [1.29] & 2.58 [3.44] \\ \cmidrule{2-4}
            & Attention ($\lambda$=0) & .497 [1.32] & 2.52 [3.25] \\
            & Attention ($\lambda$=10) & .464 [1.29] & 2.56 [3.51] \\
            & Attention ($\lambda$=20) & .463 [1.24] & {\bf{2.44 [3.20]}} \\ \midrule
            \multirow{6}{*}{HCE} & CNN+FCN~\cite{bojarski2016end} & .246 [.400] & 1.27 [1.57] \\ \cmidrule{2-4}
            & CNN+LSTM & .568 [.977] & 1.57 [2.27] \\  \cmidrule{2-4}
            & Attention ($\lambda$=0) & .334 [.766] & {\bf{1.18 [1.66]}} \\
            & Attention ($\lambda$=10) & .358 [.728] & 1.25 [1.79] \\
            & Attention ($\lambda$=20) & .373 [.724] & 1.20 [1.66] \\ \midrule
            \multirow{6}{*}{Udacity~\cite{udacity}} & CNN+FCN~\cite{bojarski2016end} & .457 [.870] & {\bf{4.12 [4.83]}} \\ \cmidrule{2-4}
            & CNN+LSTM & .481 [1.24] & 4.15 [4.93] \\ \cmidrule{2-4}
            & Attention ($\lambda$=0) & .491 [1.20] & 4.15 [4.93] \\
            & Attention ($\lambda$=10) & .489 [1.19] & 4.17 [4.96] \\
            & Attention ($\lambda$=20) & .489 [1.26] & 4.19 [4.93] \\
            \bottomrule
        \end{tabular}
        \caption{Control performance comparison in terms of mean absolute error (MAE) in degree and its standard deviation. Control accuracy is not degraded by incorporation of attention compared to an identical base CNN without attention. \textit{Abbreviation:} SD (standard deviation)}
        \label{Table:performance}
    \end{center}
\end{table}

{\small
\bibliographystyle{ieee}
\bibliography{ICCV}

\begin{thebibliography}{10}\itemsep=-1pt

\bibitem{abadi2015tensorflow}
M.~Abadi, A.~Agarwal, P.~Barham, E.~Brevdo, Z.~Chen, C.~Citro, G.~S. Corrado,
  A.~Davis, J.~Dean, M.~Devin, et~al.
\newblock Tensorflow: Large-scale machine learning on heterogeneous systems,
  2015.
\newblock {\em Software available from tensorflow. org}, 1, 2015.

\bibitem{bojarski2016visualbackprop}
M.~Bojarski, A.~Choromanska, K.~Choromanski, B.~Firner, L.~Jackel, U.~Muller,
  and K.~Zieba.
\newblock Visualbackprop: visualizing cnns for autonomous driving.
\newblock {\em arXiv preprint arXiv:1611.05418}, 2016.

\bibitem{bojarski2016end}
M.~Bojarski, D.~Del~Testa, D.~Dworakowski, B.~Firner, B.~Flepp, P.~Goyal, L.~D.
  Jackel, M.~Monfort, U.~Muller, J.~Zhang, et~al.
\newblock End to end learning for self-driving cars.
\newblock {\em arXiv preprint arXiv:1604.07316}, 2016.

\bibitem{buehler2009darpa}
M.~Buehler, K.~Iagnemma, and S.~Singh.
\newblock {\em The DARPA urban challenge: autonomous vehicles in city traffic},
  volume~56.
\newblock springer, 2009.

\bibitem{burt1983laplacian}
P.~Burt and E.~Adelson.
\newblock The laplacian pyramid as a compact image code.
\newblock {\em IEEE Transactions on communications}, 31(4):532--540, 1983.

\bibitem{chen2015deepdriving}
C.~Chen, A.~Seff, A.~Kornhauser, and J.~Xiao.
\newblock Deepdriving: Learning affordance for direct perception in autonomous
  driving.
\newblock In {\em Proceedings of the IEEE International Conference on Computer
  Vision}, pages 2722--2730, 2015.

\bibitem{chen2017brain}
S.~Chen, S.~Zhang, J.~Shang, B.~Chen, and N.~Zheng.
\newblock Brain inspired cognitive model with attention for self-driving cars.
\newblock {\em arXiv preprint arXiv:1702.05596}, 2017.

\bibitem{commaai}
Comma.ai.
\newblock {Public driving dataset}.
\newblock \url{https://github.com/commaai/research}, 2017.
\newblock [Online; accessed 07-Mar-2017].

\bibitem{ester1996density}
M.~Ester, H.-P. Kriegel, J.~Sander, X.~Xu, et~al.
\newblock A density-based algorithm for discovering clusters in large spatial
  databases with noise.
\newblock In {\em Kdd}, volume~96, pages 226--231, 1996.

\bibitem{glorot2010understanding}
X.~Glorot and Y.~Bengio.
\newblock Understanding the difficulty of training deep feedforward neural
  networks.
\newblock In {\em Aistats}, volume~9, pages 249--256, 2010.

\bibitem{hendricks2016generating}
L.~A. Hendricks, Z.~Akata, M.~Rohrbach, J.~Donahue, B.~Schiele, and T.~Darrell.
\newblock Generating visual explanations.
\newblock In {\em European Conference on Computer Vision}, pages 3--19.
  Springer, 2016.

\bibitem{hochreiter1997long}
S.~Hochreiter and J.~Schmidhuber.
\newblock Long short-term memory.
\newblock {\em Neural computation}, 9(8):1735--1780, 1997.

\bibitem{hyndman2008forecasting}
R.~Hyndman, A.~B. Koehler, J.~K. Ord, and R.~D. Snyder.
\newblock {\em Forecasting with exponential smoothing: the state space
  approach}.
\newblock Springer Science \& Business Media, 2008.

\bibitem{johnson2016densecap}
J.~Johnson, A.~Karpathy, and L.~Fei-Fei.
\newblock Densecap: Fully convolutional localization networks for dense
  captioning.
\newblock In {\em Proceedings of the IEEE Conference on Computer Vision and
  Pattern Recognition}, pages 4565--4574, 2016.

\bibitem{kingma2014adam}
D.~Kingma and J.~Ba.
\newblock Adam: A method for stochastic optimization.
\newblock {\em arXiv preprint arXiv:1412.6980}, 2014.

\bibitem{lecun2015deep}
Y.~LeCun, Y.~Bengio, and G.~Hinton.
\newblock Deep learning.
\newblock {\em Nature}, 521(7553):436--444, 2015.

\bibitem{lee2009convolutional}
H.~Lee, R.~Grosse, R.~Ranganath, and A.~Y. Ng.
\newblock Convolutional deep belief networks for scalable unsupervised learning
  of hierarchical representations.
\newblock In {\em Proceedings of the 26th annual international conference on
  machine learning}, pages 609--616. ACM, 2009.

\bibitem{levinson2011towards}
J.~Levinson, J.~Askeland, J.~Becker, J.~Dolson, D.~Held, S.~Kammel, J.~Z.
  Kolter, D.~Langer, O.~Pink, V.~Pratt, et~al.
\newblock Towards fully autonomous driving: Systems and algorithms.
\newblock In {\em Intelligent Vehicles Symposium (IV), 2011 IEEE}, pages
  163--168. IEEE, 2011.

\bibitem{pomerleau1989alvinn}
D.~A. Pomerleau.
\newblock Alvinn, an autonomous land vehicle in a neural network.
\newblock Technical report, Carnegie Mellon University, Computer Science
  Department, 1989.

\bibitem{rajamani2011vehicle}
R.~Rajamani.
\newblock {\em Vehicle dynamics and control}.
\newblock Springer Science \& Business Media, 2011.

\bibitem{sharma2015action}
S.~Sharma, R.~Kiros, and R.~Salakhutdinov.
\newblock Action recognition using visual attention.
\newblock {\em arXiv preprint arXiv:1511.04119}, 2015.

\bibitem{srivastava2014dropout}
N.~Srivastava, G.~E. Hinton, A.~Krizhevsky, I.~Sutskever, and R.~Salakhutdinov.
\newblock Dropout: a simple way to prevent neural networks from overfitting.
\newblock {\em Journal of Machine Learning Research}, 15(1):1929--1958, 2014.

\bibitem{udacity}
Udacity.
\newblock {Public driving dataset}.
\newblock \url{https://www.udacity.com/self-driving-car}, 2017.
\newblock [Online; accessed 07-Mar-2017].

\bibitem{urmson2008autonomous}
C.~Urmson, J.~Anhalt, D.~Bagnell, C.~Baker, R.~Bittner, M.~Clark, J.~Dolan,
  D.~Duggins, T.~Galatali, C.~Geyer, et~al.
\newblock Autonomous driving in urban environments: Boss and the urban
  challenge.
\newblock {\em Journal of Field Robotics}, 25(8):425--466, 2008.

\bibitem{xu2016end}
H.~Xu, Y.~Gao, F.~Yu, and T.~Darrell.
\newblock End-to-end learning of driving models from large-scale video
  datasets.
\newblock {\em arXiv preprint arXiv:1612.01079}, 2016.

\bibitem{xu2015show}
K.~Xu, J.~Ba, R.~Kiros, K.~Cho, A.~C. Courville, R.~Salakhutdinov, R.~S. Zemel,
  and Y.~Bengio.
\newblock Show, attend and tell: Neural image caption generation with visual
  attention.
\newblock In {\em ICML}, volume~14, pages 77--81, 2015.

\bibitem{zahavy2016graying}
T.~Zahavy, N.~B. Zrihem, and S.~Mannor.
\newblock Graying the black box: Understanding dqns.
\newblock {\em arXiv preprint arXiv:1602.02658}, 2016.

\bibitem{zeiler2014visualizing}
M.~D. Zeiler and R.~Fergus.
\newblock Visualizing and understanding convolutional networks.
\newblock In {\em European Conference on Computer Vision}, pages 818--833.
  Springer, 2014.

\end{thebibliography}
}
\end{document}